\pdfoutput=1
\documentclass[10pt]{article}
\usepackage[preprint]{tmlr}
\usepackage{amsmath,amssymb,amsthm}
\usepackage{booktabs}
\usepackage{graphicx}
\usepackage{hyperref}
\hypersetup{hidelinks}
\usepackage{placeins}
\usepackage{float}
\usepackage{url}
\usepackage{algorithm}
\usepackage{algpseudocode}
\usepackage{xcolor}
\usepackage{amsmath,amsfonts,bm}

\def\eqref#1{equation~\ref{#1}}
\def\1{\bm{1}}

\DeclareMathAlphabet{\mathsfit}{\encodingdefault}{\sfdefault}{m}{sl}
\SetMathAlphabet{\mathsfit}{bold}{\encodingdefault}{\sfdefault}{bx}{n}

\title{Watch Where You Linearize:\\ Predicting and Correcting Model-Merging Interference}
\author{\name Ido Nitzan Hidekel \\
      \addr School of Electrical and Computer Engineering\\
      Tel Aviv University
      \AND
      \name Dan Raviv \\
      \addr School of Electrical and Computer Engineering\\
      Tel Aviv University}

\newcommand{\tv}[1]{\tau_{#1}}

\begin{document}
\maketitle

\begin{abstract}
Model merging composes separately fine-tuned models by adding their task
vectors to a shared backbone.\footnote{Generative AI tools were used as
coding and writing assistants, as described in the AI use statement after the
Conclusion.} Interference between the tasks is usually measured only after
a candidate merge is built and evaluated, and choosing merge coefficients
repeats that build for every candidate. We show that both steps can be
replaced by a first-order quantity, the directional derivative
$J_A(\theta)\tau_B$ of task $A$'s outputs along task $B$'s vector, provided
it is read at the right point. Read at the model that receives the update,
it forecasts nearby merges per task and per example. It halves the
accuracy error of a no-change baseline and finds most of the examples whose
prediction flips. For a complete multi-task merge one read is too coarse,
and two reads along the merge path are needed. Read at the shared
pretrained initialization, the anchor implicit in task arithmetic and
tangent-space composition, the same derivative is nearly orthogonal to the
realized change in our vision settings and only partly informative in LoRA
language-model pilots. Controls tie this gap to displacement along the
receiver's own task vector rather than to parameter distance, and the gap
forms within the first few dozen fine-tuning steps. Freezing the
receiver-local responses gives a surrogate whose outputs are affine in the
merge coefficients. On it we build \textbf{LR-Merge}, a label-free
correction that fits coefficients by convex teacher matching on unlabeled
probes, without backpropagating through any merged network. LR-Merge
improves every closed-form, isotropic and sparsified base merger we test,
and its gain generally increases with the number of merged tasks. Built from pretrained
responses instead, it makes the merge worse than its base. It matches
exact teacher matching through the merged network, which takes about three
times longer to converge and is at parity at an equal budget, and one
build answers further coefficient or task-subset queries in seconds.
\end{abstract}

\section{Introduction}

\begin{figure}[t]
\centering
\includegraphics[width=0.81\textwidth]{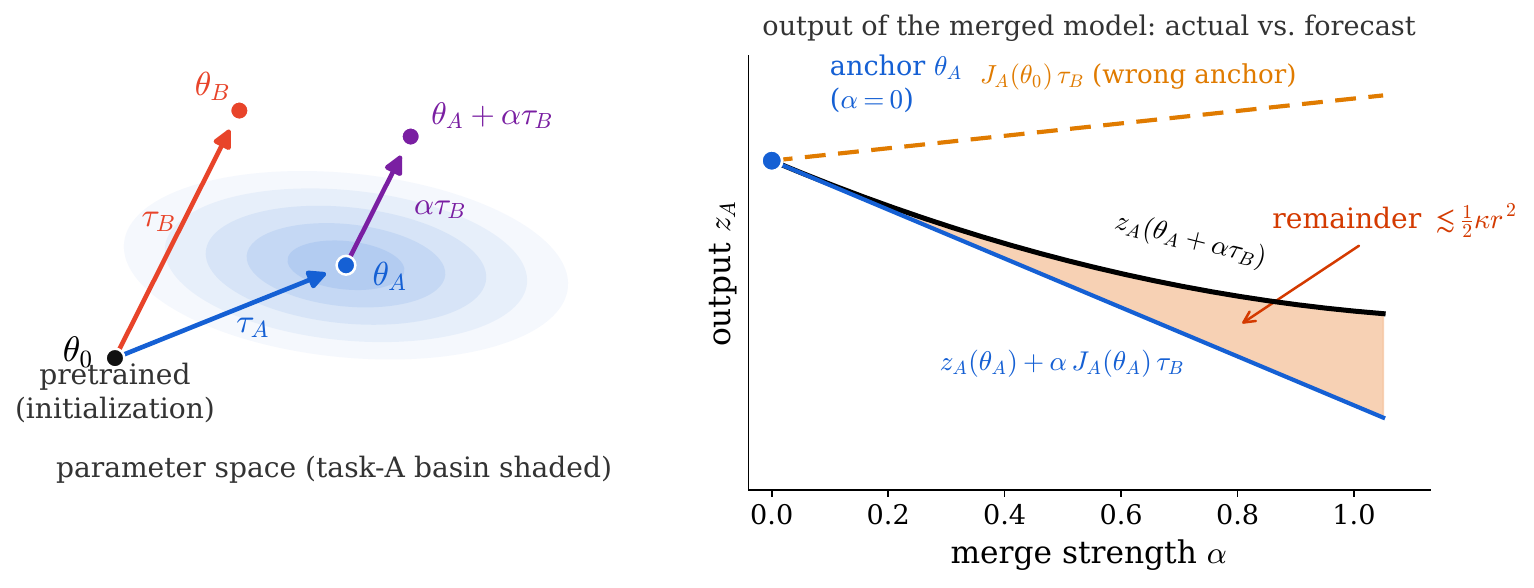}
\caption{Interference is receiver-local. $z_A(\theta)$: logits on task
$A$'s probes; $J_A{=}\partial z_A/\partial\theta$; $\tau_B$: task $B$'s
vector. Black: realized drift $z_A(\theta_A{+}\alpha\tau_B){-}z_A(\theta_A)$;
blue: forecast at the receiving model, $\alpha J_A(\theta_A)\tau_B$; orange:
forecast at the pretrained model, $\alpha J_A(\theta_0)\tau_B$, which points
in a nearly unrelated direction (formal setup in Section~\ref{sec:forecasting}).}
\label{fig:geometry}
\end{figure}

Model merging composes separately fine-tuned models into one network,
without joint training or the original data. Each model contributes its
parameter difference from the shared pretrained starting point, its
\emph{task vector} \citep{wortsman2022soups,ilharco2023task}. The updates
interfere. A vector learned for one task changes predictions on another
and costs accuracy. Closed-form mergers reduce the damage by altering the
composition analytically \citep{jin2023regmean,marczak2025isotropic,yadav2023ties}.
Adaptive mergers tune coefficients on unlabeled data, building and running
the merged network at every step \citep{yang2024adamerging}. In both cases
the effect of a merge is known only after the merged model is built and
evaluated, and every hyperparameter a pipeline tunes (coefficients,
sparsity levels, task subsets) multiplies that cost. Yet everything the
merge is made of is known in advance: the task vectors and the models
that will receive them. We ask whether the effect of adding a task vector
can be predicted before any merge is built, and whether the prediction can
replace the build inside coefficient search.

The natural object is the \emph{cross-task directional response}
$J_A(\theta)\,\tv{B}$: the first-order change task $A$'s outputs undergo
when task $B$'s vector is added at $\theta$, one Jacobian--vector product
(JVP). First-order expansion is standard. What is not settled is
\emph{where} the response should be read when task vectors come from
ordinary nonlinear fine-tuning. Task arithmetic defines every vector
relative to the pretrained $\theta_0$, tangent-space methods linearize
there \citep{ortiz2023tangent}, and kernel-regime accounts suggest the
initial linearization can persist through fine-tuning
\citep{malladi2023kernel}. All three invite reading the response at
$\theta_0$, the one point shared by all tasks. A merge, however, is
applied to a fine-tuned receiver.

We find that the anchor is decisive. Read at the receiving model,
$J_A(\theta_A)\tv{B}$ forecasts nearby merges accurately. Read at
$\theta_0$, it carries almost no usable signal in the vision settings we
study. What separates the two anchors is direction, not distance. The gap
forms at the very beginning of fine-tuning, when the network re-aims its
cross-task responses while their norm barely changes.

The finding becomes useful through a method. If receiver-local responses
are accurate near a merge, \emph{freezing} them at that merge gives a
surrogate whose outputs are affine in the merge coefficients. Coefficient
selection then becomes a convex, label-free problem: match the fine-tuned
teachers' predictions on unlabeled probes. The search never builds or
backpropagates through a candidate merged network. We call the method
\textbf{Linear-Response Merge (LR-Merge)}. It needs no labels and no joint
training. It is not data-free, since it uses a few unlabeled probes from
each task's distribution and the fine-tuned models to compute
pseudo-labels. Because a derivative is local, LR-Merge is designed as a
cheap \emph{correction} of a strong base merge rather than a merger from
scratch. The base merger supplies a good starting point, and the frozen
responses replace the search around it with convex solves whose build is
reused across queries.

\paragraph{Contributions.}
\begin{enumerate}
\item \textbf{A pre-merge interference forecast, with its range of
validity.} One JVP at the receiving model predicts, before the merge is
built, how logits drift, which examples flip, and each task's accuracy at
the strengths that coefficient search visits (Table~\ref{tab:e1}). Used to
score candidate coefficients, it ranks them as well as building and
evaluating each, and once the surrogate is built it scores a candidate
about a thousand times faster. For a
complete merge one read degrades, and two reads along the path recover
nearly all of the realized drift (Table~\ref{tab:magf}).
\item \textbf{Where the response must be read, and why.} The pretrained
anchor fails in vision. On three CLIP backbones its forecast is nearly
orthogonal to the realized drift, while the receiver-local forecast is
closely aligned with it (Tables~\ref{tab:anchors} and~\ref{tab:transfer}).
In LoRA language pilots it is weaker but keeps some signal. Anchor
controls tie the failure to displacement along the receiver's task
vector rather than to distance, and training trajectories place its onset
in the first few dozen steps (Figure~\ref{fig:onset}).
\item \textbf{LR-Merge, a cheap convex correction built on the forecast.}
It improves every closed-form, isotropic and sparsified base we test on
the canonical eight-task benchmark, the $14$- and $20$-task suites and
larger backbones, and the gain generally grows with the number of tasks
(Table~\ref{tab:main}). It matches exact teacher matching through the
merged network and AdaMerging's objective run on the same base, at a
third or less of their cost to convergence (at an equal budget, exact
matching is at parity). It beats gradient-free search at matched
wall-clock, and it amortizes, since each further coefficient query is a
convex solve of a few seconds (Section~\ref{sec:cost}).
\end{enumerate}

\section{Related Work}\label{sec:related}

\paragraph{Model merging.}
Task arithmetic \citep{ilharco2023task} adds task vectors at chosen
coefficients. One line edits the vectors before adding them: sparsifying
and resolving sign conflicts
\citep{yadav2023ties,yu2024dare,davari2024breadcrumbs,du2024pcb}, masking
parameters per task \citep{wang2024tall,qi2025tswitch,huang2024emr}, or
restructuring them in low-rank and spectral subspaces
\citep{xiong2024awd,cheng2025wudi,liu2025lore,marczak2025isotropic,zheng2025free}.
A second spends calibration data \citep{matena2022fisher,jin2023regmean,yang2024adamerging,lee2026adarank,wei2025apgd,yang2024surgery}.
Both learn what a composition did by building it, whereas LR-Merge corrects
their outputs through a frozen first-order surrogate. Output-space projection merging \citep{evans2026osp} also works
in output space, but constructs new layer-wise updates, whereas we keep the task
vectors fixed and optimize only their coefficients.

\paragraph{Linearized task arithmetic.}
Tangent-space fine-tuning makes linearity true by construction, training
within a linearization fixed at $\theta_0$
\citep{ortiz2023tangent,liu2023tangent,tang2024parameter}, and
\citet{ortiz2023tangent} show that post-hoc linearization of a nonlinear
fine-tune at $\theta_0$ loses accuracy. Regularizers make vectors
interfere less or behave more linearly: Jacobian-product penalties
\citep[$\tau$Jp,][]{yoshida2025tjp}, curvature-based disentanglement
\citep{porrello2026tak}, and distillation of linearized behavior
\citep{sommariva2026delta}. These change how checkpoints are trained. We
take ordinary checkpoints as given and ask post hoc where their response
must be read. For a fully linearized model the two anchors coincide by
construction, but on our six tasks such fine-tunes lose several points of
single-task accuracy (Table~\ref{tab:supp}).

\paragraph{Predicting the effect of a merge.}
\citet{zhou2026demystifying} predict a scalar, whether a merge will
succeed, from model-pair metrics, and SyMerge \citep{symerge2026} relates
cross-task performance to merging quality and, like LR-Merge, uses expert
pseudo-labels, but backpropagates through the merged network. MAP
\citep{li2025map} amortizes coefficient search by fitting quadratic
surrogates of task metrics to a set of evaluated merges, for Pareto-front
estimation. LR-Merge instead linearizes model outputs with receiver-local
JVPs and solves a convex teacher-matching problem, without evaluating each trial
coefficient setting through the full merged network. Our forecast is per example and
per coefficient. It answers the scalar question as a by-product, since one
frozen build ranks all $57$ task subsets of our six-task suite by real
merged accuracy with high rank correlation. \citet{daheim2024model} bound
merging error through gradient mismatch. On our pairs only the
receiver-side term of their bound ranks interference, and less well than
our label-free forecast (Table~\ref{tab:supp}). Influence
functions \citep{koh2017influence,grosse2023studying} and tangent overlap
in continual learning \citep{doan2021ntk,evron2022catastrophic,hidekel2026catastrophic}
estimate effects of parameter changes more broadly, and \citet{zhou2024ctl}
observe cross-task linearity of features between endpoints.

\section{Predicting Merge Interference}
\label{sec:forecasting}

\paragraph{Setup.} Unless stated otherwise, measurements use CLIP
ViT-B/32 \citep{radford2021clip} on six tasks (MNIST, SVHN, GTSRB, EuroSAT,
DTD, CIFAR-10), fine-tuned independently under the standard task-vector
protocol \citep{ilharco2023task}. Each task has $256$ unlabeled probes and
an index-disjoint $1000$-example evaluation split, and every measurement is
replicated over six independently fine-tuned checkpoint sets. This suite uses CIFAR-10 where
the canonical suite uses Stanford Cars, only because the Cars host was
unavailable when the study began. The forecast and the correction were
repeated on the canonical eight tasks (MNIST, SVHN, GTSRB, EuroSAT, DTD,
Cars, RESISC45, SUN397), and those are the numbers to compare with prior
work. Table~\ref{tab:suites} (Appendix~\ref{app:repro}) lists every suite
and its checkpoints.

\paragraph{The forecasting problem.} Fine-tuning task $t\in\{1,\ldots,T\}$
from $\theta_0\in\mathbb{R}^P$ gives $\theta_t$ and task vector
$\tv{t}:=\theta_t-\theta_0$. For a target task $A$ with probes
$X_A=\{x_{A,i}\}_{i=1}^{n_A}$, let $z_A(\theta)\in\mathbb{R}^{n_AC_A}$ be the
stacked logits ($C_A$ classes, fixed heads absorbed) and
$J_A(\theta):=\partial z_A(\theta)/\partial\theta$. Adding task $B$'s vector
to the model fine-tuned on $A$ displaces its outputs by
$\Delta z_{A\leftarrow B}(\alpha):=z_A(\theta_A+\alpha\tv{B})-z_A(\theta_A)$,
and first-order expansion at the receiver gives
\begin{equation}
z_A(\theta_A+\alpha\tv{B})=z_A(\theta_A)+\alpha\,J_A(\theta_A)\tv{B}+R_{A,B}(\alpha),
\label{eq:pairwise-taylor}
\end{equation}
which defines the \emph{receiver-local forecast}
$\widehat{\Delta z}_{A\leftarrow B}(\alpha):=\alpha\,J_A(\theta_A)\tv{B}$: one
JVP, no merge built. Appendix~\ref{app:bounds} bounds $R_{A,B}$ by
curvature along the merge segment.

\paragraph{Metrics.} For each ordered pair $A\leftarrow B$ and strength
$\alpha$, on the $n{=}256$ probes of $A$: \emph{drift cosine} is the cosine
between the flattened $nC_A$-dimensional $\widehat{\Delta z}$ and
$\Delta z$. \emph{Damage $\rho$} is the Spearman correlation over probes
between forecast and realized change of the true-class logit.
\emph{Acc.\ MAE} is the absolute difference between forecast accuracy
($\arg\max$ of $z_A(\theta_A)+\widehat{\Delta z}$) and realized merged
accuracy. A \emph{flip} is a probe whose $\arg\max$ differs from that of
$\theta_A$, pooled over pairs. Drift cosine, damage $\rho$ and MAE are
averaged over the $30$ ordered pairs, then over three probe seeds. Labels
enter only damage $\rho$ and accuracy. The \emph{no-change} reference
predicts the receiver's own accuracy.

\begin{table}[t]
\centering
\small
\caption{Pairwise forecast, $30$ ordered pairs, checkpoint set $0$, three
probe seeds (other sets in Table~\ref{tab:e1seeds}). Flips are rare at
$\alpha{=}0.2$ ($1.0\%$ of probe--pair cases, and $17$ of $30$ pairs have
none). Rate-matched label-free low-margin and random flaggers reach
precision $0.57$ and $0.06$ ($0.42$, $0.11$ at $\alpha{=}0.4$). At
$\alpha{=}0.4$ the per-pair accuracy drop correlates at $0.75$ with the
forecast and at $0.12$ with task-vector cosine.}
\label{tab:e1}
\scalebox{0.9}{\begin{tabular}{lcccccc}
\toprule
 & \multicolumn{2}{c}{acc.\ MAE (pp)} & & & & \\
\cmidrule(lr){2-3}
$\alpha$ & forecast & no change & flip recall & flip prec. & damage $\rho$ & drift cos\\
\midrule
0.2 & 0.13 & 0.26 & 0.82 & 0.79 & 0.917 & 0.946\\
0.4 & 0.30 & 0.60 & 0.75 & 0.68 & 0.824 & 0.905\\
\bottomrule
\end{tabular}
}
\end{table}

\paragraph{Pairwise forecasts.} Table~\ref{tab:e1} reports the forecast at
the strengths coefficient search visits. It predicts per-task accuracy to
within a few tenths of a point, recovers about four in five flipping
examples at similar precision, and orders which examples are hurt.
Against the no-change reference it halves the accuracy error at every
strength we test. Its larger added value is per example, where no-change
predicts nothing. The forecast replicates on all six checkpoint sets and
on all $56$ pairs of the canonical eight tasks (Table~\ref{tab:e1seeds}).
Scores computed from the task vectors alone do not substitute for it.
Task-vector cosine barely tracks the per-pair accuracy drop, which the
forecast tracks closely (Table~\ref{tab:supp}). When the receiver is
itself a scaled model $\theta_0+\beta\tau_A$, the forecast read at that
receiver stays accurate and the $\theta_0$ forecast stays poor
(Table~\ref{tab:beta}), so the anchor is whichever model receives the
update. Used to score candidate coefficients, the forecast ranks them as
well as building and evaluating each (Appendix~\ref{app:extra}).

\begin{table}[t]
\centering
\small
\caption{Forecasting complete merges ($s{=}1$) produced by five base
mergers: six receivers, mean over three checkpoint sets. One JVP is too
coarse at this distance, and two along the path are accurate. At $s{=}0.2$
one JVP at $\theta_A$ reaches cosine $0.90$--$0.98$ for every merger, against
$-0.14$ to $0.17$ at $\theta_0$.}
\label{tab:magf}
\scalebox{0.9}{\begin{tabular}{lcccccccc}
\toprule
& \multicolumn{3}{c}{one JVP at $\theta_A$} & \multicolumn{3}{c}{two JVPs on the path (midpoint)} & \multicolumn{2}{c}{one JVP at $\theta_0$}\\
\cmidrule(lr){2-4}\cmidrule(lr){5-7}\cmidrule(lr){8-9}
merger & cos & damage $\rho$ & acc MAE & cos & damage $\rho$ & acc MAE & cos & damage $\rho$\\
\midrule
RegMean & 0.756 & 0.668 & 4.0 & 0.978 & 0.973 & 1.1 & -0.209 & -0.004\\
Iso-C & 0.793 & 0.537 & 3.5 & 0.989 & 0.978 & 2.1 & 0.160 & 0.104\\
Iso-CTS & 0.788 & 0.470 & 5.3 & 0.986 & 0.971 & 2.8 & 0.044 & 0.039\\
TIES & 0.674 & 0.479 & 11.8 & 0.959 & 0.950 & 3.2 & 0.191 & 0.243\\
Task arithmetic & 0.721 & 0.509 & 11.3 & 0.977 & 0.964 & 3.1 & 0.243 & 0.207\\
\bottomrule
\end{tabular}
}
\end{table}

\paragraph{Complete merges: the forecast is local.} The forecast applies to
any merger as a black box. A merge $\theta_M$ displaces the receiver by
$\theta_M-\theta_A$, and $J_A(\theta_A)(\theta_M-\theta_A)$ forecasts that
displacement. A fifth of the way to the merge it is well aligned for all
five mergers, while the $\theta_0$ forecast is close to orthogonal
(Table~\ref{tab:supp}). At the merge itself a single read degrades and
misses per-task accuracy by several points (Table~\ref{tab:magf}), so one
JVP does not replace evaluating a complete merge. Reading the response at
two points along the path, the midpoint rule for
$\int_0^1 J_A(\gamma(s))\Delta\,ds$ (Appendix~\ref{app:bounds}), recovers
nearly all of the realized drift. This shapes the design of LR-Merge,
which reads the response at the current merge and uses it only for nearby
corrections.

\section{Why the Forecast Must Be Receiver-Local}
\label{sec:geometry}

Both forecasts are first-order expansions. The $\theta_A$ forecast is a
Taylor expansion at the point where the update is applied. The $\theta_0$
forecast transports a derivative from another point, and its error is the
change of $J_A\tau_B$ along the anchor displacement, a curvature term.
The question for merging is
whether the error grows with distance, so that any nearby
anchor will do, or depends on direction, so that the anchor inherited
from task arithmetic is the wrong one. We find the latter, locate when it
forms, and show in Section~\ref{sec:wrong-anchor} that it decides whether
a correction helps or harms.

\begin{table}[t]
\centering
\small
\caption{Forecast drift cosine by anchor: $30$ ordered pairs at
$\alpha{=}0.2$, mean$\pm$sd over six checkpoint sets, min $=$ worst pair.
Last column: $|\langle d,\tau_A\rangle|/\lVert\tau_A\rVert^2$ for anchor
displacement $d$ from $\theta_A$. This displacement orders the nine anchors
(Spearman $-0.98$, $p{=}3\times10^{-5}$), and distance from $\theta_A$ does
not ($-0.16$, $p{=}0.70$).}
\label{tab:anchors}
\resizebox{0.9\linewidth}{!}{\begin{tabular}{llcccc}
\toprule
anchor & what it tests & mean cos & min cos & dist.\ from $\theta_A$ & disp.\ along $\tau_A$\\
\midrule
$\theta_A$ (receiving model) & the forecast as used & 0.949$\pm$0.008 & 0.296 & $0$ & $0$\\
isotropic, matched distance & is distance the cause? & 0.945$\pm$0.009 & 0.315 & $\lVert\tv{A}\rVert$ & 0.00\\
isotropic, $1.41\times$ distance & farther still random & 0.942$\pm$0.009 & -- & $1.41\,\lVert\tv{A}\rVert$ & 0.00\\
span of other task vectors & is the direction structured? & 0.665$\pm$0.036 & $-0.320$ & $\lVert\tv{A}\rVert$ & 0.07\\
toward $\theta_C$, matched distance & third model, confound removed & 0.461$\pm$0.014 & $-0.007$ & $\lVert\tv{A}\rVert$ & 0.67\\
isotropic tilted $45^\circ$ to $-\tau_A$ & direction dose--response & 0.352$\pm$0.018 & -- & $\lVert\tv{A}\rVert$ & 0.71\\
$\theta_C$, a third fine-tuned model & does any fine-tuned point work? & 0.349$\pm$0.014 & $-0.650$ & $1.41\,\lVert\tv{A}\rVert$ & 0.94\\
$\theta_0$ (pretrained) & the natural alternative & 0.122$\pm$0.011 & $-0.719$ & $\lVert\tv{A}\rVert$ & 1.00\\
$\theta_0$ $+$ isotropic kick & is the point $\theta_0$ special? & 0.116$\pm$0.022 & -- & $1.41\,\lVert\tv{A}\rVert$ & 1.00\\
\bottomrule
\end{tabular}
}
\end{table}

\paragraph{Direction, not distance.} Table~\ref{tab:anchors} moves the
anchor. Isotropic anchors as far from $\theta_A$ as $\theta_0$, or farther,
forecast as well as the receiver. Tilting an isotropic anchor $45^\circ$
toward $-\tau_A$ cuts the drift cosine by almost two thirds. A third
fine-tuned model does no better, and a full-size isotropic kick away from
$\theta_0$ does not revive it. Displacement along the receiver's own task
vector orders all nine anchors almost perfectly, while distance does not
order them (Table~\ref{tab:supp}). Displacement along other tasks' vectors
also costs, but less. The span-of-others anchor barely moves along
$\tau_A$ and still loses about a third of its alignment, so this
coordinate is the dominant factor but not a complete characterization. We
therefore use ``receiver-local'' only for the rule \emph{read the response
at the model being updated}. That some other anchors also work is an
empirical observation about our settings.

\begin{figure}[t]
\centering
\includegraphics[width=0.83\linewidth]{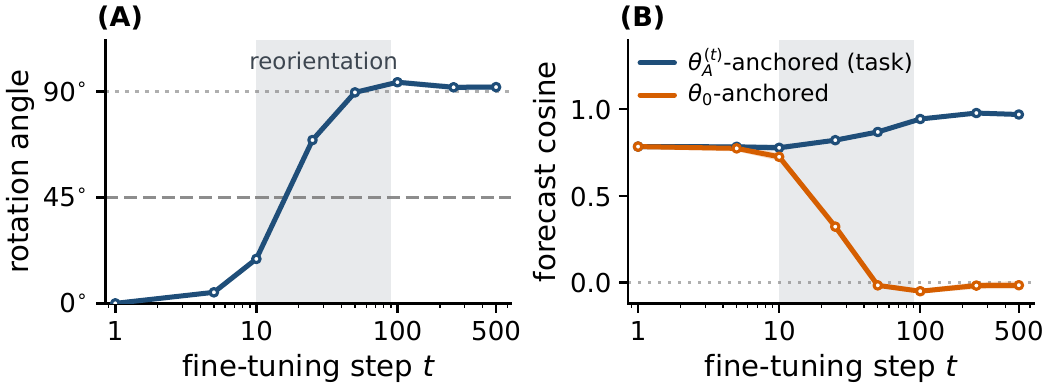}
\caption{Cross-task responses re-aim at the start of fine-tuning.
(A)~Angle between $J_A(\theta_A^{(t)})\tv{B}$ and $J_A(\theta_0)\tv{B}$ along
the receiver's fine-tuning. (B)~The $\theta_0$-anchored forecast collapses
over the same window, while the forecast at the current checkpoint keeps rising.
Receivers SVHN and GTSRB, five sources each, $\tau_B$ fixed at the source's
final checkpoint, $\alpha{=}0.2$, no labels. Lines are means over pairs and
three fine-tuning seeds, and the band is the across-seed spread ($t\le500$, seed $0$
beyond).}
\label{fig:onset}
\end{figure}

\paragraph{An early re-aiming.} We track
$r_{A\leftarrow B}(t):=J_A(\theta_A^{(t)})\tv{B}$ at checkpoints
$t\in\{1,5,10,25,50,\ldots,2000\}$ of the receiver's fine-tuning
(Figure~\ref{fig:onset}). Its angle to the initialization response passes
$45^\circ$ within roughly the first twenty steps and saturates by step
$50$, when the weights have covered only a fifth of their final
displacement. The $\theta_0$ forecast collapses over the same window. Two
further receivers replicate the early crossing at different plateau
heights. The response rotates rather than shrinks. Its norm stays almost
constant, and the initial and final responses keep most of their energy
in the same low-dimensional output subspace while becoming nearly
orthogonal. Norm-matched random directions rotate on the same schedule, so
the onset is generic to the evolving Jacobian. What is pair-specific is
the plateau, where cross-task angles spread about three times more widely
than random ones (Table~\ref{tab:supp}, Appendix~\ref{app:anchor}).

\section{LR-Merge: Freezing the Local Response}
\label{sec:method}

Forecasting one candidate is not yet a merging method, because coefficient
selection compares many nearby candidates. LR-Merge reads the responses
once at a merge, freezes them, and searches coefficients inside the
resulting surrogate.

\paragraph{Coefficients and reference model.} A base merger (RegMean,
Iso-C, TIES, \ldots) outputs $\theta_{\mathrm{base}}$. LR-Merge searches over
\emph{total} coefficients $\boldsymbol\alpha\in\mathbb{R}^M$ on the task
vectors added to it,
$\theta(\boldsymbol\alpha):=\theta_{\mathrm{base}}+\sum_{t=1}^{T}\alpha_t\tv{t}$,
starting from $\boldsymbol\alpha{=}0$ (the base itself). In the
\emph{grouped} variant each block-restricted piece $\Pi_g\tv{t}$ ($\Pi_g$
zeroes parameters outside group $g$) receives its own coefficient, so
$M{=}TG$, with $G{=}13$ groups on ViT-B (one per transformer block plus
embedding/projection) and $25$ on ViT-L/14. The reported corrections use
the grouped variant, and everything below reads unchanged with $t$
replaced by $(t,g)$.

\paragraph{One round.} Round $r$ holds coefficients $\mathbf a^{(r)}$ and
(i) forms the anchor $\bar\theta^{(r)}:=\theta(\mathbf a^{(r)})$; (ii) reads
and freezes the responses $D^{(r)}_{A,t}:=J_A(\bar\theta^{(r)})\tv{t}$ for
every target $A$ and source $t$, giving the affine surrogate
\begin{equation}
\widehat z^{(r)}_A(\boldsymbol\alpha):=z_A(\bar\theta^{(r)})+\sum_{t}\bigl(\alpha_t-a^{(r)}_t\bigr)D^{(r)}_{A,t};
\label{eq:affine-surrogate}
\end{equation}
(iii) fits $\boldsymbol\alpha$ by minimizing
\begin{equation}
F_r(\boldsymbol\alpha):=\frac{1}{T}\sum_{A=1}^{T}\frac{1}{n_A}\sum_{i=1}^{n_A}
\ell_{\mathrm{CE}}\!\left(\widehat z^{(r)}_A(\boldsymbol\alpha)_{i,\cdot},\,\widehat y_{A,i}\right)
+\lambda\lVert\boldsymbol\alpha-\mathbf a^{(r)}\rVert_2^2,
\label{eq:lrmerge-objective}
\end{equation}
where $\widehat y_{A,i}:=\arg\max_c z_A(\theta_A)_{i,c}$ are pseudo-labels
from the fine-tuned teacher on unlabeled probes (so LR-Merge needs the
teachers and unlabeled task-distribution probes, but no labels); (iv) applies the clipped
coefficients $\mathbf a^{(r+1)}:=\mathrm{clip}_{[l,u]}(\boldsymbol\alpha^{(r)}_{\mathrm{Adam}})$;
and (v) moves the anchor there and rebuilds the responses for the next
round. We use two rounds throughout (Algorithm~\ref{alg:lrmerge}).

\begin{algorithm}[t]
\caption{LR-Merge}
\label{alg:lrmerge}
\small
\begin{algorithmic}[1]
\Require task vectors $\{\tv{t}\}$, teachers $\{\theta_A\}$, unlabeled probes $\{X_A\}$, base $\theta_{\mathrm{base}}$, box $[l,u]$, $\lambda$, rounds $R$
\State $\widehat y_{A,i}\gets\arg\max_{c}\,z_A(\theta_A)_{i,c}\ \ \forall A,i$; \ $\mathbf a^{(0)}\gets 0$ \Comment{pseudo-labels, once}
\For{$r=0,\ldots,R-1$}
  \State $\bar\theta^{(r)}\gets\theta_{\mathrm{base}}+\sum_{t}a^{(r)}_t\,\tv{t}$ \Comment{anchor}
  \State $D^{(r)}_{A,t}\gets J_A(\bar\theta^{(r)})\,\tv{t}\ \ \forall A,t$ \Comment{one JVP each; frozen}
  \State $\boldsymbol\alpha^{(r)}_{\mathrm{Adam}}\gets\textsc{Adam}\bigl[F_r;\ \text{init }\mathbf a^{(r)}\bigr]$ \Comment{objective (\ref{eq:lrmerge-objective}) on surrogate (\ref{eq:affine-surrogate})}
  \State $\mathbf a^{(r+1)}\gets\mathrm{clip}_{[l,u]}\bigl(\boldsymbol\alpha^{(r)}_{\mathrm{Adam}}\bigr)$
\EndFor
\State \Return $\theta_{\mathrm{base}}+\sum_{t}a^{(R)}_t\,\tv{t}$
\end{algorithmic}
\end{algorithm}

\paragraph{Why it is cheap, and convex.} Once frozen, every candidate's
logits are linear combinations of stored vectors, so no merged network is
built, run, or backpropagated through during the search. Because the
surrogate logits are affine in $\boldsymbol\alpha$, $F_r$ is convex and,
with the ridge, $2\lambda$-strongly convex with a unique minimizer
(Appendix~\ref{app:bounds}). We minimize it with a fixed Adam schedule.
Clipping after an unconstrained solve is not a projected method. In every
deployed run the coefficients stay strictly inside the box
(Table~\ref{tab:supp}), so clipping leaves the optimizer's output
unchanged. We match hard
pseudo-labels with cross-entropy, as does the exact teacher-matching
baseline of Section~\ref{sec:cost}. This works as well as KL to the
teacher's soft labels and clearly better than regressing teacher logits
(Table~\ref{tab:supp}). The ridge keeps each move small, and the second
round re-reads the geometry after the first. This is how the method
respects the locality of Section~\ref{sec:forecasting}, and at the
deployed anchors the forecast stays well aligned (Appendix~\ref{app:anchor}).

\paragraph{Cost, and a low-rank build.} Reading the responses costs $T^2G$
JVPs per anchor, which grows quadratically with the number of tasks. A
rank-$k$ output projection removes that growth. For each target, the
$TG$ projected responses are inner products of the task vectors with $k$
vector--Jacobian products that depend on the target alone
(Appendix~\ref{app:bounds}), so the build costs $O(Tk)$ passes. Rank $16$
matches the exact build's accuracy (Table~\ref{tab:controls}) and is used
for the $14$- and $20$-task suites. A deployed correction keeps the probes,
their cached pseudo-labels (not the teachers), the task vectors, and the
response tensors, about $10$\,MB per anchor for six tasks
(Table~\ref{tab:cost}).

\section{Experiments}
\label{sec:experiments}

All wall-clocks are end-to-end on one GPU with identical probe budgets
(RTX 8000 unless noted; Appendix~\ref{app:repro}).

\subsection{LR-Merge Improves Closed-Form, Isotropic and Sparsified Mergers}
\label{sec:refine}

\begin{table}[t]
\centering
\small
\caption{LR-Merge as a correction of base mergers (average accuracy, \%).
Rows with one checkpoint set are pilots without across-set spread. The
correction cost excludes the base merger (Iso-C: $54$\,s). Iso-C,
Iso-CTS and TIES select their global scale by labeled probe accuracy, while
RegMean, DOGE and LR-Merge use no labels. Per-set
values and the eight-task suite with CIFAR-10 in Appendix~\ref{app:seeds}.}
\label{tab:main}
\resizebox{0.9\linewidth}{!}{\begin{tabular}{llllll}
\toprule
suite (backbone) & checkpoints & base merger & base & $+$LR-Merge & $\Delta$\\
\midrule
canonical eight (B/32) & ours, 3 sets & RegMean & 80.70$\pm$0.04 & \textbf{82.73$\pm$0.12} & $+$2.03$\pm$0.08\\
 & & Iso-C & 83.90$\pm$0.22 & \textbf{85.36$\pm$0.03} & $+$1.46$\pm$0.19\\
 & & Iso-CTS & 84.23$\pm$0.07 & \textbf{86.08$\pm$0.06} & $+$1.85$\pm$0.09\\
\midrule
six-task (B/32) & ours, 6 sets & RegMean & 87.49$\pm$0.22 & \textbf{90.58$\pm$0.20} & $+$3.10$\pm$0.37\\
 & & Iso-C & 88.78$\pm$0.35 & \textbf{91.28$\pm$0.27} & $+$2.50$\pm$0.31\\
 & & Iso-CTS & 88.15$\pm$0.38 & \textbf{91.49$\pm$0.23} & $+$3.34$\pm$0.49\\
 & & TIES & 77.06$\pm$0.66 & \textbf{86.08$\pm$0.45} & $+$9.02$\pm$0.34\\
 & ours, 3 sets & DOGE (adaptive) & 88.73$\pm$0.18 & \textbf{89.53$\pm$0.32} & $+$0.79$\pm$0.20\\
\midrule
14 tasks (B/32) & released, 1 set & RegMean & 76.94 & \textbf{80.51} & $+$3.57\\
 & & Iso-CTS & 79.38 & \textbf{82.37} & $+$2.99\\
20 tasks (B/32) & released, 1 set & RegMean & 71.09 & \textbf{77.34} & $+$6.25\\
 & & Iso-CTS & 73.35 & \textbf{79.29} & $+$5.94\\
\midrule
six-task (B/16) & ours, 1 set & RegMean / Iso-C / Iso-CTS & 89.27 / 92.18 / 92.37 & 91.87 / 93.00 / 93.35 & $+$2.60 / $+$0.82 / $+$0.98\\
six-task (L/14) & ours, 1 set & RegMean / Iso-C / Iso-CTS & 92.83 / 93.53 / 94.57 & 94.18 / 94.65 / 95.05 & $+$1.35 / $+$1.12 / $+$0.48\\
\midrule
canonical eight (B/32) & released, 1 run & AdaRank (adaptive) & 86.56 & 86.34 & $-$0.22\\
\bottomrule
\end{tabular}
}
\end{table}

Table~\ref{tab:main} collects the correction results. On the canonical
eight-task benchmark LR-Merge improves RegMean, Iso-C and Iso-CTS on every
checkpoint set. On the six-task suite it improves all four bases on every
set, and it gives a smaller but significant gain on the recent adaptive
merger DOGE \citep{wei2025apgd}. The gain generally grows with the number
of merged tasks. Over a nested ordering of two to eight tasks it is
negligible at two tasks, exceeds three points from six tasks on, and is
significant for every $T\ge5$ (Table~\ref{tab:tsweep}). With released
$14$- and $20$-task checkpoints \citep{wang2024tall} it is larger still and
largest at $20$ tasks, concentrated on tasks where the bases collapse
(EMNIST, KMNIST). The same direction holds on ViT-B/16 and ViT-L/14.

AdaRank \citep{lee2026adarank} adapts its merge with $10$k entropy steps on
the test images and is already close to its teachers. LR-Merge still
lowers its held-out teacher-matching loss, but no label-free setting we
tried moves its accuracy by more than a few tenths of a point
(Appendix~\ref{app:extra}). Given the same $256$ unlabeled probes instead
of the test pool, AdaRank falls below RegMean with LR-Merge
(Table~\ref{tab:supp}). LR-Merge pulls a merge toward its teachers, so it
helps in proportion to how far the base is from them.

\paragraph{The anchor is part of the algorithm.}
\label{sec:wrong-anchor}
The wrong-anchor ablation changes one thing. It reads the responses at
$\theta_0$, $D_{A,t}=J_A(\theta_0)\tv{t}$, instead of at the current anchor
$\bar\theta^{(r)}$, and keeps the objective, probes, pseudo-labels, solver,
rounds and wall-clock. The RegMean correction then loses about seven points
on every checkpoint set and ends below the uncorrected base, with most of
the damage on SVHN and GTSRB (Table~\ref{tab:controls}; per set in
Table~\ref{tab:anchorabl}). In standalone selection the same swap costs
about as much. A wrong anchor thus turns a helpful first-order correction into a
harmful one.

\subsection{Cost and Comparisons on Equal Footing}
\label{sec:cost}

\begin{table}[t]
\centering
\small
\caption{Correcting the same RegMean base by different optimizers. Paired
differences are (row $-$ LR-Merge) over the listed checkpoint sets.}
\label{tab:controls}
\resizebox{0.9\linewidth}{!}{\begin{tabular}{lcccc}
\toprule
correction of the same RegMean base & $\Delta$ accuracy & paired vs.\ LR-Merge & wall-clock (s) & sets\\
\midrule
\multicolumn{5}{l}{\emph{six-task suite}}\\
exact teacher matching, converged & $+$3.24$\pm$0.23 & $+0.14\pm0.24$ & 3447$\pm$78 & 6\\
exact teacher matching, stopped at LR-Merge's budget & $+$3.18$\pm$0.30 & $+0.08\pm0.14$ & ${\approx}900$ & 6\\
AdaMerging objective on the same base, converged & $+$3.13$\pm$0.52 & $+0.03\pm0.22$ & 3432$\pm$32 & 6\\
coordinate descent on the same objective, matched time & $+$2.87$\pm$0.27 & $-0.41\pm0.28$ & 964 & 3\\
\textbf{LR-Merge}, exact responses & $+$3.10$\pm$0.37 & -- & 952$\pm$15 & 6\\
\textbf{LR-Merge}, rank-16 sketch & $+$3.18$\pm$0.33 & $+0.08\pm0.12$ & \textbf{704$\pm$6} & 6\\
LR-Merge with responses read at $\theta_0$ (wrong anchor) & $-$3.78 & $-6.88\pm0.76$ & ${\approx}952$ & 6\\
\midrule
\multicolumn{5}{l}{\emph{canonical eight-task benchmark}}\\
exact teacher matching & $+$1.75 & $-0.28\pm0.27$ & 5290 & 3\\
AdaMerging objective on the same base, converged & $+$0.91 & $-1.12\pm0.21$ & 5087 & 3\\
\textbf{LR-Merge} & $+$2.03$\pm$0.08 & -- & \textbf{1685} & 3\\
\bottomrule
\end{tabular}
}
\end{table}

\paragraph{Same base, same objective.} Table~\ref{tab:controls} isolates
what the surrogate buys by holding the base fixed. Exact teacher matching
backpropagates the same objective through the real merged network. On the
six-task suite, run to convergence, it is within noise of LR-Merge at
about $3.6$ times the wall-clock, and stopped at LR-Merge's budget it is at
parity. On the canonical eight tasks it trails LR-Merge at three times the
cost. AdaMerging's entropy objective run to convergence on the same base
ties on the six-task suite and trails by about a point on the canonical
eight. From the stronger Iso-C and Iso-CTS bases LR-Merge is also ahead of
it, in under a third of its wall-clock (Table~\ref{tab:supp}).
Gradient-free search over the same coefficients at matched wall-clock does
not substitute. Coordinate descent trails slightly, and random search over
several hundred evaluated candidates trails by three to five points
(Table~\ref{tab:supp}). The surrogate therefore reaches the accuracy of
exact optimization from cheap gradients, with no stopping step to choose.
Its optimum is close to the exact one, since continuing exact optimization
from it adds only a fifth of a point. In standalone selection,
output-space projection merging \citep{evans2026osp} is about ten points
less accurate than task-wise LR-Merge and about $25$ times slower
(Table~\ref{tab:standalone}).

\begin{figure}[t]
\centering
\includegraphics[width=0.9\textwidth]{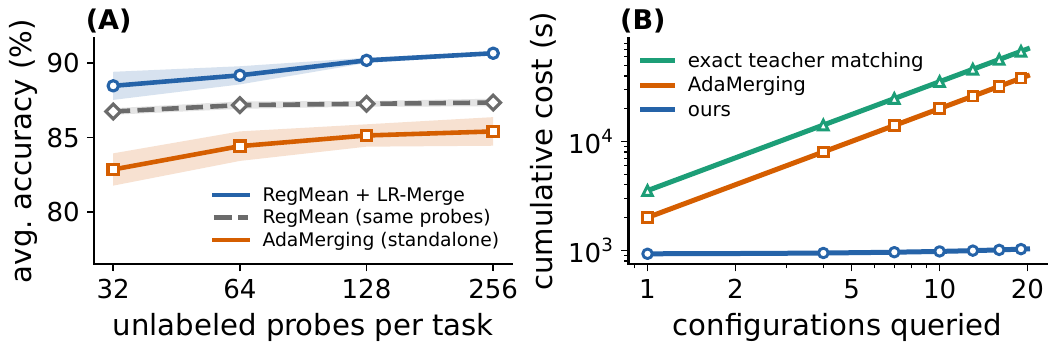}
\caption{(A)~Accuracy vs unlabeled probe budget on identical pools:
RegMean built from the same probes (dashed), LR-Merge on that RegMean, and
standalone layer-wise AdaMerging (values in Table~\ref{tab:probes}).
(B)~Cumulative cost of repeated coefficient queries. At twenty queries
LR-Merge is $39$--$69\times$ cheaper. Mean$\pm$sd over three checkpoint
sets.}
\label{fig:advantage}
\end{figure}

\paragraph{Few probes.} Figure~\ref{fig:advantage}A compares pipelines at
matched probe budgets. RegMean is built from $b$ probes and corrected by
LR-Merge with the same $b$, and standalone AdaMerging uses the same pool.
At $32$ probes per task LR-Merge still improves its RegMean base, and it
beats AdaMerging given eight times as many probes at about $26$ times less
compute (Table~\ref{tab:probes}). The arms start from different points, and
Table~\ref{tab:controls} gives the equal-footing comparison. Part of the
gap is the objective. With $32$ probes over DTD's $47$ classes,
entropy-based selection falls far behind teacher matching
(Table~\ref{tab:supp}).

\paragraph{Reuse.} The build is a cache. Once built, it answers each
further coefficient query (a new ridge, clamp, or task subset) with a
convex solve of a few seconds. Twenty queries then cost about $40$ times
less than per-configuration AdaMerging and about $70$ times less than
per-configuration exact matching (Figure~\ref{fig:advantage}B). A fast
AdaMerging implementation (500 steps, bf16) is quicker for a single
configuration but about five points less accurate on the canonical eight
tasks (Table~\ref{tab:supp}). On an L40S a build takes minutes on ViT-B/32
and under two and a half hours on ViT-L/14, and a solve takes under two
seconds (Table~\ref{tab:cost}). Re-anchoring is a new build. When a sixth
task arrives at a deployed five-task merge, computing only its responses
and re-solving is several times cheaper than a full rebuild and keeps the
five deployed tasks. The arriving task pays for the stale anchor, and a
re-anchor recovers part of that loss (Table~\ref{tab:arrival}).

\subsection{Transfer to Other Backbones and to Language Models}
\label{sec:transfer}

\begin{table}[t]
\centering
\small
\caption{Forecast drift cosine at the receiver vs at $\theta_0$ (vision at
$\alpha{=}0.2$; language in teacher-forced logit space, LoRA fine-tunes).
Language rows are pilots.}
\label{tab:transfer}
\scalebox{0.9}{\begin{tabular}{lcc}
\toprule
arm & cos @ receiver & cos @ $\theta_0$\\
\midrule
CLIP ViT-B/32 (six sets) & 0.949 & 0.122\\
CLIP ViT-B/32, canonical eight ($56$ pairs) & 0.975 & 0.180\\
CLIP ViT-B/16 & 0.966 & 0.152\\
CLIP ViT-L/14 & 0.924 & 0.090\\
\midrule
RoBERTa-base/LoRA (3 GLUE tasks) & 0.77 & 0.11\\
Pythia-410M/LoRA (5 tasks) & 0.92 & 0.23--0.68\\
Pythia-1.4B/LoRA (2 pairs) & 0.97--0.98 & 0.53--0.54\\
\bottomrule
\end{tabular}
}
\end{table}

The anchor gap holds on larger CLIP backbones and narrows, without
closing, on LoRA language models (Table~\ref{tab:transfer}). There the
pretrained anchor keeps partial signal, consistent with kernel-regime
descriptions of fine-tuning at scale \citep{malladi2023kernel}. As a
correction of a task-arithmetic merge of five Pythia-410M LoRA fine-tunes,
LR-Merge lowers held-out teacher-forced NLL and raises token accuracy in
about a minute, matching exact teacher matching through the merged network
at the same budget (Table~\ref{tab:llm-cost}). These are pilots on one
fine-tune with teacher-forced outputs. The forecast ranks teacher-forced
loss changes but not which free generations degrade
(Table~\ref{tab:supp}), and we make no claim about generation.

\section{Limitations}\label{sec:limitations}

One read of the forecast is accurate for nearby candidates, not for a
complete merge (Table~\ref{tab:magf}). A one-JVP curvature estimate ranks
where it fails (Table~\ref{tab:supp}) but is no certificate. The anchor
controls are empirical, not causal, and checkpoints trained to stay linear
($\tau$Jp, TAK, DELTA) were not measured. LR-Merge assumes one shared
initialization and architecture and needs a few tens of unlabeled probes
per task. It matches rather than exceeds exact teacher matching at matched
compute, and it does not help mergers already close to their teachers. The
larger-backbone, $14$/$20$-task, task-arrival (one arriving task) and
language results each rest on one checkpoint set or fine-tune.

\section{Conclusion}\label{sec:conclusion}

A single directional derivative, read at the model receiving a task
vector, forecasts the interference that vector causes on nearby merges,
per task and per example. Read at the pretrained model it does not, in
vision, because fine-tuning re-aims the cross-task response within its
first steps. Frozen at a merge, the same derivative makes coefficient
selection a cheap convex problem. LR-Merge corrects closed-form, isotropic
and sparsified merges, most strongly when many tasks are merged. It is as
accurate as exact teacher matching run to convergence at a third of its
cost, and it answers later queries in seconds. For the nonlinear
fine-tunes studied here, what a task vector does depends strongly on the
model that receives it, so watch where you linearize.

\subsection*{Reproducibility statement}
Section~\ref{sec:forecasting} and Appendix~\ref{app:repro} specify models,
data splits, probe/evaluation disjointness, fine-tuning and selection
hyperparameters, and hardware, and Appendix~\ref{app:seeds} reports per-set
results. Code, per-experiment scripts, stored result files and the
fine-tuned checkpoint sets will be released with the paper.

\subsection*{AI use statement}
We used generative AI tools as coding and writing assistants for
generating and refactoring experimental and analysis code, for checking
the proofs in Appendix~\ref{app:bounds}, and for refining text. All
AI-assisted code was reviewed and its outputs verified by the authors
(re-running key configurations and validating tables and figures against
the result files), and all AI-assisted text was reviewed and edited by the
authors. We did not use generative AI for research ideation.

\bibliography{references}
\bibliographystyle{tmlr}

\clearpage
\appendix

\section{Additional Quantitative Results}
\label{app:supp}

Table~\ref{tab:supp} reports the measurements behind statements in the
main text that no other table covers, grouped by section.

\begin{table}[H]
\centering
\small
\caption{Additional quantitative results, grouped by the main-text section
they support. ViT-B/32, six-task suite unless noted.}
\label{tab:supp}
\scalebox{0.9}{\begin{tabular}{p{0.52\textwidth}p{0.40\textwidth}}
\toprule
measurement & value\\
\midrule
\multicolumn{2}{l}{\emph{Section~\ref{sec:related}: related approaches}}\\
single-task accuracy, fully linearized vs ordinary fine-tunes (six tasks) & $87.5$ vs $93.8$\\
ranking all $57$ task subsets by merged accuracy, one frozen build & Spearman $0.88\pm0.03$\\
ranking pairwise interference: gradient-mismatch bound, receiver-side term / full mismatch / our forecast & $0.66$ / $0.34$ / $0.75$\\
\midrule
\multicolumn{2}{l}{\emph{Section~\ref{sec:forecasting}: forecast}}\\
flip precision at $\alpha{=}0.2$, 95\% CI & $0.79$, $[0.72,0.91]$\\
acc.\ MAE at $\alpha{=}1.0$, forecast vs no change & $2.38$ vs $4.70$\\
\midrule
\multicolumn{2}{l}{\emph{Section~\ref{sec:geometry}: anchor}}\\
$45^\circ$ crossing step, median (range) over $30$ pair$\times$seed curves & $16$ ($12.9$--$22.3$)\\
weight displacement at saturation (step $50$), fraction of final & ${\approx}20\%$\\
response norm ratio, final vs initialization (median) & $0.97$\\
cosine of the two responses, sharing a top-$32$ output subspace & $0.11$\\
spread of cross-task angles at $t{=}500$, task vs random directions & sd $27^\circ$ vs $9^\circ$\\
\midrule
\multicolumn{2}{l}{\emph{Section~\ref{sec:method}: method}}\\
smallest distance of a solved coefficient to the box & $0.48$\\
standalone task-wise selection by objective (set $0$): hard-label CE / KL to soft labels / logit MSE / closed-form ridge & $80.7$ / $80.6$ / $75.0$ / $69.3$\\
\midrule
\multicolumn{2}{l}{\emph{Section~\ref{sec:experiments}: experiments}}\\
gain vs number of tasks, Spearman over $21$ $(T,\text{set})$ points & $0.92$\\
AdaRank: held-out teacher-matching loss reduction / largest accuracy gain & $5$--$17\%$ / $0.3$\\
AdaRank with $256$ probes vs RegMean $+$ LR-Merge (canonical eight) & $82.00$ vs $83.59$\\
wrong-anchor correction, change on SVHN / GTSRB & $-24.5$ / $-7.9$\\
LR-Merge minus converged AdaMerging objective, Iso-C / Iso-CTS base (three sets) & $+0.33$ / $+0.46$\\
wall-clock of those runs, LR-Merge vs AdaMerging objective & ${\approx}950$\,s vs $3357$--$5610$\,s\\
random search at matched wall-clock (${\sim}580$ candidates), minus LR-Merge, six / grouped coefficients & $-2.6$ / $-5.1$\\
$300$ further exact teacher-matching steps from the LR-Merge solution & $+0.20\pm0.07$\\
DTD at $32$ probes, entropy vs teacher matching & $47.8$ vs $66.0$\\
reuse: build / per-query solve (RTX 8000) & $926$\,s / $5.4$\,s\\
twenty queries: LR-Merge / per-configuration AdaMerging / exact matching & $1033$ / $40{,}133$ / $70{,}867$\,s\\
fast AdaMerging (500 steps, bf16) vs RegMean $+$ LR-Merge, canonical eight & $78.68$ in $123$\,s vs $83.59$ in $774$\,s\\
Pythia-410M: Spearman with teacher-forced loss change / with free-generation degradation & $0.77$ / $0.02$\\
\midrule
\multicolumn{2}{l}{\emph{Section~\ref{sec:limitations}: limitations}}\\
one-JVP curvature estimate, Spearman with forecast failure & $0.991$\\
\bottomrule
\end{tabular}
}
\end{table}

\section{Bounds, Convexity and the Low-Rank Identity}
\label{app:bounds}

\paragraph{Pairwise remainder.} For $v:=\alpha\tv{B}$ with
$\ell:=\lVert v\rVert_2$ and $u:=v/\ell$, let $h(s):=z_A(\theta_A+s\,u)$,
$s\in[0,\ell]$, so $\ell\,h'(0)=\alpha J_A(\theta_A)\tv{B}$ is the forecast.
Taylor's theorem with integral remainder gives the forecast error exactly,
and bounding the integrand by the worst output acceleration
$\kappa_A:=\sup_{s\in[0,\ell]}\lVert h''(s)\rVert_2$ gives
\begin{equation}
z_A(\theta_A{+}v)-z_A(\theta_A)-J_A(\theta_A)v=\int_0^{\ell}(\ell-s)\,h''(s)\,\mathrm{d}s,
\qquad
\bigl\lVert z_A(\theta_A{+}v)-z_A(\theta_A)-J_A(\theta_A)v\bigr\rVert_2\le\tfrac12\kappa_A\,\ell^2 .
\label{eq:app-remainder-bound}
\end{equation}

\paragraph{Merge paths.} For a reference $\vartheta$, candidate
$\theta^\star$, $\Delta:=\theta^\star-\vartheta$ and
$\gamma(s):=\vartheta+s\Delta$,
$z_A(\theta^\star)-z_A(\vartheta)=\int_0^1J_A(\gamma(s))\Delta\,\mathrm{d}s$.
The $K$-node composite midpoint forecast
$\frac1K\sum_{k=1}^{K}J_A(\gamma(\tfrac{2k-1}{2K}))\Delta$ has error at most
$\frac{1}{24K^2}\sup_s\lVert\varphi'''(s)\rVert_2=O(\lVert\Delta\rVert_2^3/K^2)$
for $\varphi(s):=z_A(\gamma(s))$ three times continuously differentiable.
On six-way merges at $\alpha{=}0.3$, going from one node to four raises
drift cosine from $0.878$ to $0.995$ and cuts accuracy MAE from $3.97$ to
$0.35$ points over six checkpoint sets.

\paragraph{Convexity.} For fixed target $A$ and probe $i$ the surrogate
logits are affine,
$\bigl(\widehat z^{(r)}_A(\boldsymbol\alpha)_{i,\cdot}\bigr)^{\top}=b_{A,i}+M_{A,i}(\boldsymbol\alpha-\mathbf a^{(r)})$,
with $b_{A,i}:=(z_A(\bar\theta^{(r)})_{i,\cdot})^\top$ and $j$-th column of
$M_{A,i}$ equal to $\bigl((J_A(\bar\theta^{(r)})v_j)_{i,\cdot}\bigr)^\top$,
$v_j$ the $j$-th correction direction ($\tv{t}$ or $\Pi_g\tv{t}$).
Cross-entropy has logit Hessian $\mathrm{Diag}(p)-pp^\top\succeq0$, so each
per-example term has Hessian
$M_{A,i}^\top[\mathrm{Diag}(p_{A,i})-p_{A,i}p_{A,i}^\top]M_{A,i}\succeq0$, and
the ridge adds $2\lambda I$: $F_r$ is $2\lambda$-strongly convex with a
unique minimizer. The clamp is applied after the solve and is not part of
the convex problem.

\paragraph{Low-rank identity.} For an orthonormal $U_k$ spanning the
leading output directions,
$U_k^{\top}J_A(\bar\theta)[\Pi_g\tv{t}]=\bigl(J_A(\bar\theta)^{\top}U_k\bigr)^{\!\top}[\Pi_g\tv{t}]$,
where $J_A(\bar\theta)^{\top}U_k$ is $k$ vector--Jacobian products for target
$A$ alone. $U_k$ is found by a randomized range finder with $q$ power
iterations and oversampling $p$, for $T(2q{+}2)(k{+}p)$ passes in all
($q{=}1$, $p{=}5$).

\section{Receiver-Locality: Additional Measurements}
\label{app:anchor}

\paragraph{Deployed anchors.} LR-Merge reads responses at a base merge far
from every receiver ($73\pm17$ times $\lVert\tv{A}\rVert$ for RegMean) but
nearly orthogonal to the task-vector span (projection fraction $0.015$,
per-unit-distance projection on $\tv{A}$ $-0.009$), and $99.9\%$ of its
displacement from $\theta_0$ lies in the first two transformer blocks,
where RegMean's least-squares solve is ill-conditioned. Measured directly
with forecast and realized drift both around $\bar\theta$, the drift
cosine is $0.858\pm0.006$ at $\delta{=}0.2$ and $0.909\pm0.017$ at the
typical solved correction magnitude. Interpolating
Table~\ref{tab:anchors} linearly in per-unit-distance projection, between
the isotropic row ($0.945$ at $0$) and the span row ($0.665$ at $0.07$),
gives $0.909$ at $\bar\theta$'s $0.009$, matching the measurement.

\paragraph{Scope controls.} A class-split control (two disjoint halves of
CIFAR-100) and a from-scratch ResNet pilot place the anchor effect in
fine-tuning of pretrained encoders, not in distribution shift.

\paragraph{Onset details.} Per-seed $45^\circ$ crossings on the primary
receivers are $16.19/15.82/16.15$. Until saturation the angle tracks
weight displacement (Spearman $+0.94$ for $t\le50$). Norm-matched random
Gaussian directions cross at $t\approx17.6$ against $\approx17.9$ for
task vectors. At $t{=}500$ cross-task angles spread to sd $27^\circ$ against
$\pm9^\circ$ for random directions (Levene $p<10^{-3}$). MNIST and EuroSAT
receivers cross at $t\approx8.9$ and $11.7$ and plateau near $97^\circ$
and $75^\circ$, and their $\theta_0$ forecast collapses from $0.83$ to $0.07$
while the current-anchor forecast reaches $0.975$. Along the straight
segment $\theta_0+s\tv{A}$ the response rotates smoothly with $s$, so the
checkpoint result is not an artifact of training time.

\begin{table}[H]
\centering
\small
\caption{Drift cosine for a scaled receiver $\theta_0+\beta\tau_A$: forecast
at the actual receiver / at $\theta_A$ / at $\theta_0$, mean over $30$ pairs
and three probe seeds (seed spread ${\le}0.02$). At $\beta{=}1$ the
receiver is $\theta_A$.}
\label{tab:beta}
\scalebox{0.9}{\begin{tabular}{lcccc}
\toprule
$\beta$ & $\alpha{=}0.2$ & $\alpha{=}0.4$ & $\alpha{=}0.7$ & $\alpha{=}1.0$\\
\midrule
0.3 & \textbf{0.83} / 0.42 / 0.29 & \textbf{0.69} / 0.43 / 0.23 & \textbf{0.51} / 0.45 / 0.15 & 0.37 / \textbf{0.45} / 0.05\\
0.5 & \textbf{0.91} / 0.66 / 0.21 & \textbf{0.82} / 0.63 / 0.20 & \textbf{0.68} / 0.59 / 0.16 & \textbf{0.56} / 0.55 / 0.10\\
0.7 & \textbf{0.94} / 0.83 / 0.15 & \textbf{0.88} / 0.79 / 0.15 & \textbf{0.77} / 0.73 / 0.14 & \textbf{0.66} / 0.65 / 0.12\\
1.0 & \textbf{0.95} / -- / 0.11 & \textbf{0.91} / -- / 0.12 & \textbf{0.83} / -- / 0.11 & \textbf{0.74} / -- / 0.11\\
\bottomrule
\end{tabular}
}
\end{table}

\section{Per-Set Results}
\label{app:seeds}

\begin{table}[H]
\centering
\small
\caption{Forecast and anchor gap over six independently fine-tuned
checkpoint sets, $30$ ordered pairs each at $\alpha{=}0.2$. Last row: the
canonical eight tasks, all $56$ ordered pairs of one checkpoint set.}
\label{tab:e1seeds}
\scalebox{0.9}{\begin{tabular}{lcccc}
\toprule
checkpoint set & cos @ $\theta_A$ & cos @ $\theta_0$ & damage $\rho$ & acc.\ MAE (pp)\\
\midrule
$0$ & 0.946 & 0.124 & 0.916 & 0.13\\
$1$ & 0.954 & 0.126 & 0.925 & 0.17\\
$2$ & 0.958 & 0.107 & 0.927 & 0.16\\
$3$ & 0.946 & 0.109 & 0.916 & 0.09\\
$4$ & 0.953 & 0.132 & 0.930 & 0.13\\
$5$ & 0.936 & 0.133 & 0.916 & 0.10\\
\midrule
mean $\pm$ sd & \textbf{0.949$\pm$0.008} & \textbf{0.122$\pm$0.011} & 0.922$\pm$0.006 & 0.13$\pm$0.03\\
\midrule
canonical eight, $56$ pairs & 0.975 & 0.180 & 0.946 & 0.22\\
\bottomrule
\end{tabular}
}
\end{table}

\begin{table}[H]
\centering
\small
\caption{Wrong-anchor ablation (six-task, six checkpoint sets): responses
read at the current anchor $\bar\theta^{(r)}$ vs at $\theta_0$, all else
fixed, in standalone selection and in the RegMean correction. The
correction drop has $t(5){=}22.2$.}
\label{tab:anchorabl}
\scalebox{0.9}{\begin{tabular}{lccccccc}
\toprule
responses read at & s0 & s1 & s2 & s3 & s4 & s5 & mean $\pm$ sd\\
\midrule
\multicolumn{8}{l}{\emph{standalone selection}}\\
current anchor $\bar\theta^{(r)}$ & 80.65 & 80.18 & 80.72 & 79.98 & 80.83 & 80.00 & \textbf{80.39$\pm$0.38}\\
pretrained $\theta_0$ & 73.77 & 73.83 & 73.80 & 74.00 & 74.65 & 73.65 & 73.95$\pm$0.36\\
paired drop & 6.88 & 6.35 & 6.92 & 5.98 & 6.18 & 6.35 & \textbf{6.44$\pm$0.38}\\
\midrule
\multicolumn{8}{l}{\emph{correction of RegMean (base $87.49\pm0.22$)}}\\
current anchor $\bar\theta^{(r)}$ & 90.53 & 90.68 & 90.73 & 90.23 & 90.80 & 90.52 & \textbf{90.58$\pm$0.20}\\
pretrained $\theta_0$ & 84.12 & 83.67 & 84.07 & 84.45 & 82.90 & 83.03 & 83.71$\pm$0.63\\
paired drop & 6.41 & 7.01 & 6.66 & 5.78 & 7.90 & 7.49 & \textbf{6.88$\pm$0.76}\\
\bottomrule
\end{tabular}
}
\end{table}

\begin{table}[H]
\centering
\small
\caption{Eight-task suite with CIFAR-10 in place of Cars, six checkpoint
sets.}
\label{tab:refine8}
\scalebox{0.9}{\begin{tabular}{lccccc}
\toprule
base merger (8-task) & base & $+$ours & $\Delta$ & sets & cost (s)\\
\midrule
RegMean & 83.10$\pm$0.25 & \textbf{86.19$\pm$0.20} & $+3.09\pm0.34$ & 6 & 1685\\
Iso-C & 85.92$\pm$0.22 & \textbf{88.28$\pm$0.19} & $+2.36\pm0.15$ & 6 & 1700\\
Iso-CTS & 86.18$\pm$0.46 & \textbf{88.66$\pm$0.22} & $+2.47\pm0.52$ & 6 & 1688\\
TIES & 73.62$\pm$0.44 & \textbf{80.62$\pm$0.65} & $+7.00\pm0.31$ & 6 & 1693\\
\bottomrule
\end{tabular}
}
\end{table}

\begin{table}[H]
\centering
\small
\caption{Average accuracy vs number of merged tasks $T$ on the eight-task
suite (tasks added in the order MNIST, SVHN, GTSRB, EuroSAT, DTD, CIFAR-10,
RESISC45, SUN397), mean$\pm$sd over three
checkpoint sets; task arithmetic at $\alpha{=}0.3$, and $p$ from a paired test
of the correction against its base.}
\label{tab:tsweep}
\scalebox{0.9}{\begin{tabular}{lccccc}
\toprule
$T$ & task arithmetic & RegMean base & RegMean $+$ ours & gain & $p$\\
\midrule
2 & 90.50$\pm$0.39 & 98.23$\pm$0.21 & 98.28$\pm$0.15 & $+0.05\pm0.17$ & 0.67\\
3 & 86.06$\pm$0.48 & 97.09$\pm$0.16 & 97.41$\pm$0.19 & $+0.32\pm0.24$ & 0.15\\
4 & 81.73$\pm$0.25 & 96.18$\pm$0.21 & 97.03$\pm$0.21 & $+0.85\pm0.35$ & 0.052\\
5 & 75.23$\pm$0.12 & 88.82$\pm$0.20 & 91.05$\pm$0.15 & $+2.23\pm0.34$ & 0.0077\\
6 & 75.80$\pm$0.15 & 87.37$\pm$0.25 & 90.65$\pm$0.10 & $+3.28\pm0.35$ & 0.0037\\
7 & 75.77$\pm$0.29 & 86.53$\pm$0.10 & 89.98$\pm$0.15 & $+3.45\pm0.25$ & 0.0018\\
8 & 73.28$\pm$0.45 & 82.98$\pm$0.19 & 86.29$\pm$0.08 & $+3.31\pm0.23$ & 0.0015\\
\bottomrule
\end{tabular}
}
\end{table}

\begin{table}[H]
\centering
\small
\caption{Per-budget values behind Figure~\ref{fig:advantage}A, checkpoint
sets $0$--$2$. RegMean@$b$ is built from the same $b$ probes that LR-Merge
uses, and AdaMerging is standalone. Building RegMean adds under $5$\,s.}
\label{tab:probes}
\scalebox{0.9}{\begin{tabular}{lccccc}
\toprule
probes & RegMean@$b$ & + LR-Merge & AdaMerging & gap & LR / Ada (s)\\
\midrule
32 & 86.75$\pm$0.22 & \textbf{88.47$\pm$0.94} & 82.86$\pm$1.08 & $+$5.61 & 130 / 2378\\
64 & 87.18$\pm$0.25 & \textbf{89.18$\pm$0.61} & 84.42$\pm$0.99 & $+$4.76 & 246 / 3416\\
128 & 87.27$\pm$0.19 & \textbf{90.18$\pm$0.18} & 85.14$\pm$0.75 & $+$5.04 & 480 / 3435\\
256 & 87.35$\pm$0.26 & \textbf{90.66$\pm$0.11} & 85.41$\pm$0.96 & $+$5.25 & 943 / 3437\\
\bottomrule
\end{tabular}
}
\end{table}

\FloatBarrier
\section{Additional Experiments}
\label{app:extra}

\begin{table}[H]
\centering
\small
\caption{Cost of one response build and one convex solve, six tasks,
$256$ probes per task, one L40S GPU. ViT-L/14 uses a JVP batch of $8$.}
\label{tab:cost}
\scalebox{0.9}{\begin{tabular}{lcccccc}
\toprule
backbone & params & groups & build (s) & solve (s) & peak GB & cache MB\\
\midrule
ViT-B/32 & 88M & 13 & 201 & 1.5 & 6.5 & 10\\
ViT-B/16 & 86M & 13 & 838 & 1.6 & 6.8 & 10\\
ViT-L/14 & 304M & 25 & 8704 & 1.4 & 19.8 & 19\\
\bottomrule
\end{tabular}
}
\end{table}

\begin{table}[H]
\centering
\small
\caption{A sixth task arrives at a deployed five-task corrected merge
(three checkpoint sets). $^\dagger$Six-task standalone run of
Table~\ref{tab:standalone}, whose per-task CIFAR-10 accuracy is not
reported.}
\label{tab:arrival}
\scalebox{0.9}{\begin{tabular}{lcccc}
\toprule
 & accuracy (6 tasks) & deployed 5 tasks & new task (CIFAR-10) & cost (s)\\
\midrule
incremental (new JVPs $+$ re-solve) & 89.54$\pm$0.34 & 90.7 & 83.7$\pm$2.4 & \textbf{146$\pm$1}\\
$+$ surrogate re-anchor & 90.09$\pm$0.33 & 90.6 & 87.5$\pm$2.0 & 613\\
full rebuild (base $+$ correction) & 90.65$\pm$0.10 & 89.9 & 94.5 & 964\\
AdaMerging from scratch$^\dagger$ & 85.2$\pm$0.8 & -- & -- & 2007\\
\bottomrule
\end{tabular}
}
\end{table}

\begin{table}[H]
\centering
\small
\caption{Standalone selection (six-task, from scratch). LR-Merge matches
the optimization-based methods at a fraction of their cost. Closed-form
Iso-C is stronger standalone, which is why correction is the intended
use. With six re-anchoring rounds LR-Merge reaches $86.17\pm0.61$.}
\label{tab:standalone}
\scalebox{0.9}{\begin{tabular}{lcc}
\toprule
method (six-task, from scratch) & avg & sel.\ cost (s)\\
\midrule
Iso-C (closed-form) & \textbf{88.9$\pm$0.3} & 54\\
AdaMerging (task-wise) & 80.1$\pm$0.5 & 1185\\
AdaMerging (layer-wise) & 85.2$\pm$0.8 & 2007\\
Exact teacher matching (layer-wise) & 85.6 & 2160\\
OSP, per-row, all layers (reimpl.; set 0) & 70.8 & 2161\\
OSP, final MLP only (reimpl.; set 0) & 65.9 & 49\\
LR-Merge (task-wise) & 80.4$\pm$0.4 & \textbf{84}\\
LR-Merge (layer-wise) & 85.4$\pm$0.5 & 933\\
\bottomrule
\end{tabular}
}
\end{table}

\paragraph{Ranking candidate coefficients.} One build of $36$ JVPs at a
merged model ($c{=}0.3$), then $60$ random coefficient vectors within
$\pm0.1$ of it, scored by the forecast and, separately, by building and
evaluating each on the same $256$ probes (checkpoint set $0$). Against
held-out accuracy the forecast ranks them at Spearman $0.947$ (direct
evaluation $0.955$), with the same top-1 regret ($0.28$ points), at
$1.3$\,ms instead of $1.56$\,s per candidate. Anchored at $\theta_A$ over a
wider box, which the candidates move away from, it drops to $0.50$--$0.67$.

\paragraph{AdaRank.} Canonical eight tasks, released checkpoints
\citep{ilharco2023task}, one L40S GPU, $1000$ evaluation images per task.
Ported into our evaluation code with its official CART configuration
($10$k entropy steps on the test pool, bf16), AdaRank reaches $86.56$ at
its final step and $87.69$ at its best step (chosen by us with labels),
below its reported $89.2$. LR-Merge on top gives $86.56\to86.34$ and
$87.69\to87.26$ at $256$ probes. A label-free search over $56$ settings
(trust region or proximal penalty around AdaRank's coefficients, one or
two rounds, per-task gating, $256$ or $512$ probes), selected by held-out
teacher-matching loss, gains at most $0.3$ points.

\paragraph{DOGE.} Adaptive projective gradient descent \citep{wei2025apgd}
with AdaMerging-style coefficients on the same $256$ probes per task
reaches $88.73\pm0.18$, and LR-Merge raises it to $89.53\pm0.32$. With $32$
probes per task the gain disappears ($-0.81\pm0.98$, not significant).

\paragraph{Full test sets.} Re-evaluating the six-task models of
checkpoint set $0$ on the complete official test sets changes each
average by at most $0.65$ points and no ordering (RegMean $87.74$, with
LR-Merge $90.99$, Iso-C $88.76$, with LR-Merge $91.35$).

\paragraph{Language-model arm.} RoBERTa-base with rank-$8$ LoRA on three
GLUE tasks, and Pythia-410M/1.4B with rank-$16$ LoRA. Forecasts use
held-out teacher-forced token blocks and target the full
token-by-vocabulary logit displacement. At $1.4$B, one JVP takes
$14.6$\,s and $20$\,GB against $6.2$\,s and $5.5$\,GB for a forward pass
(about $2.4$ forward passes), and exact teacher matching needs $1227$\,s to
converge.
RoBERTa uses SST-2, MRPC and QNLI, with LoRA rank $8$, $\alpha{=}16$ on the query
and value projections, AdamW, learning rate $10^{-4}$, $1000$ steps of batch
$32$, maximum length $128$. Pythia-410M: math (GSM8K), code (MBPP), French
(opus\_books en--fr), question answering (SQuAD v1.1) and topic (AG News),
with LoRA rank $16$, $\alpha{=}32$ on the attention and MLP projections, AdamW,
learning rate $10^{-4}$ with cosine decay, $800$ steps of batch $8$,
$256$-token packed blocks. Forecasts use $64$ held-out probe blocks per
task, and each correction draw fits on $16$ blocks and evaluates on $16$
disjoint blocks. Pythia-1.4B: the two ordered pairs math$\leftrightarrow$French at
$\alpha{=}0.3$.

\begin{table}[H]
\centering
\small
\caption{Pythia-410M/LoRA correction, mean$\pm$sd over three
probe/optimizer draws (one LoRA fine-tune, teacher-forced held-out
metrics). Both methods beat the base ($p{=}0.006$ NLL, $p{=}0.0015$
token accuracy), and oracle stopping steps differ across draws
($500/50/100$).}
\label{tab:llm-cost}
\scalebox{0.9}{\begin{tabular}{lccc}
\toprule
arm (Pythia-410M/LoRA, five tasks) & held-out NLL & top-1 token acc.\ (\%) & wall (s)\\
\midrule
base merge (TA $\alpha{=}0.3$) & 2.1154$\pm$0.0208 & 57.10$\pm$0.49 & --\\
\textbf{LR-Merge (ours)} & 2.0895$\pm$0.0225 & 57.83$\pm$0.51 & \textbf{78}\\
exact TM at ours' budget & 2.0935$\pm$0.0213 & 57.89$\pm$0.52 & $\leq$78\\
exact TM, oracle best step & 2.0924$\pm$0.0222 & 57.88$\pm$0.53 & --\\
exact TM, final ($1000$ steps) & 2.1059$\pm$0.0228 & 57.85$\pm$0.48 & 1227\\
\bottomrule
\end{tabular}
}
\end{table}

\FloatBarrier
\section{Reproducibility Details}
\label{app:repro}

\begin{table}[H]
\centering
\small
\caption{Task suites and checkpoints (ViT-B/32 unless noted).}
\label{tab:suites}
\scalebox{0.9}{\begin{tabular}{p{0.14\textwidth}p{0.44\textwidth}p{0.34\textwidth}}
\toprule
suite & tasks & checkpoints; used in \\
\midrule
six-task & MNIST, SVHN, GTSRB, EuroSAT, DTD, \mbox{CIFAR-10} & ours, six sets (one on B/16, L/14); \mbox{Sections 3--6} \\
eight-task & six-task $+$ RESISC45, SUN397 & ours, six sets; Tables~\ref{tab:refine8}, \ref{tab:tsweep} \\
canonical eight & MNIST, SVHN, GTSRB, EuroSAT, DTD, Cars, RESISC45, SUN397 & ours, three sets: Tables~\ref{tab:main}, \ref{tab:controls}; released \citep{ilharco2023task}: AdaRank, fast AdaMerging \\
14 tasks & canonical eight $+$ CIFAR-100, STL-10, Flowers102, Oxford-IIIT Pet, PCAM, FER2013 & released \citep{wang2024tall}, one set \\
20 tasks & 14 tasks $+$ EMNIST, CIFAR-10, Food-101, Fashion-MNIST, Rendered SST-2, KMNIST & released \citep{wang2024tall}, one set \\
\bottomrule
\end{tabular}}
\end{table}

\paragraph{Splits.} Each task uses $256$ unlabeled probes for JVPs,
forecasting and coefficient selection, and an index-disjoint
$1000$-example evaluation split, and probe seeds vary independently of
fine-tuning seeds. Released $14$/$20$-task checkpoints match their
published accuracies within $0.11$ points. Standard deviations over six
checkpoint sets are sample deviations, while three-set controls report
population deviations.

\paragraph{Fine-tuning.} AdamW, learning rate $10^{-5}$, weight decay
$0.1$, batch size $128$, $2000$ iterations, cosine decay, frozen zero-shot
text heads, and the non-QuickGELU OpenAI-weight CLIP variant of
\texttt{open\_clip}. Tangent-space fine-tunes use the same settings.

\paragraph{JVPs.} Forward-mode automatic differentiation in evaluation
mode, with no full Jacobians or per-example gradients. The math
scaled-dot-product-attention backend is used because the fused backend
does not support forward-mode derivatives. Software: PyTorch 2.8.0 (CUDA
12.8) and \texttt{open\_clip} 3.3.0. Merged parameters are evaluated with
\texttt{functional\_call}, without building new modules.

\paragraph{LR-Merge.} Ridge $\lambda=0.01$, Adam with learning rate
$0.01$ for $800$ steps (task-wise and grouped correction) or $1500$ steps
(grouped standalone selection). Initial coefficients $0$ for corrections
and $0.3$ for standalone selection, box $[-1,1.5]$ for corrections and
$[0,1.5]$ for selection, two rounds. Within $[-0.5,1.0]$ to $[-1,3]$ the
box changes no coefficient and no accuracy, while a sign-constrained $[0,1.5]$
box clips $46\%$ of coordinates and costs $0.85\pm0.37$ points. Reported
cost covers both builds, the solver and re-anchoring, and pseudo-labels are
computed once.

\paragraph{Base mergers.} \emph{Task arithmetic}: $\theta_0+\alpha\sum_t\tau_t$
with $\alpha{=}0.3$. \emph{RegMean} \citep{jin2023regmean}: per linear layer,
$W=(\sum_tW_tG_t)(\sum_tG_t)^{-1}$ with input Gram matrices $G_t$ computed on
each task's $256$ probes through its own fine-tuned model, off-diagonal
reduction $\gamma{=}0.9$ and ridge $10^{-2}$. Other parameters (LayerNorm,
embeddings, biases) are averaged. Because \texttt{nn.MultiheadAttention}
calls its output projection functionally, a forward hook on that module
never fires, so we route the attention context through the module (verified
equivalent to $10^{-5}$) so that the $12$ output projections also receive
the RegMean solve, and the reported RegMean bases use this fix. \emph{Iso-C} and \emph{Iso-CTS} \citep{marczak2025isotropic}: ports of
the official code (isotropic spectrum per 2-D weight, Iso-CTS with
common-subspace fraction $0.8$), with the global scale chosen from
$\{0.1,0.2,\dots,3.0\}$ by average accuracy on the labeled probes
(early stopping after three non-improving steps), as in the authors'
protocol but with our probes as the validation split. \emph{TIES}
\citep{yadav2023ties}: keep the top $20\%$ of $|\tau_t|$ per task, elect
signs, average agreeing entries, with the scale chosen from $\{0.4,0.7,1.0,1.3\}$
by labeled probe accuracy, as for Iso-C and Iso-CTS. \emph{DOGE}
\citep{wei2025apgd}: adaptive projective gradient descent with
AdaMerging-style coefficients on the same $256$ probes per task.
\emph{AdaRank} \citep{lee2026adarank}: official CART configuration
(Appendix~\ref{app:extra}).

\paragraph{Comparison optimizers (Table~\ref{tab:controls}).} All use the
same base, the same grouped $T\times G$ coefficients on the task vectors
($G{=}13$), the same $256$ unlabeled probes and pseudo-labels, and the same
$1000$-example evaluation split. \emph{Exact teacher matching}:
cross-entropy to the pseudo-labels through the real merged network, Adam
(learning rate $10^{-3}$) from zero coefficients, one batch of $64$ probes
per task per step, $2000$ steps, accuracy logged at
$100,200,300,500,1000,1500,2000$ steps (the matched-budget point is read off
this curve). \emph{AdaMerging objective on the same base}: identical, with
prediction entropy in place of cross-entropy. \emph{Random search}:
coefficient vectors drawn uniformly in the correction box $[-1,1.5]$, each
scored by the teacher-matching objective on the real merged network, best
kept, until the wall-clock budget ($952$\,s, LR-Merge's) is spent.
\emph{Coordinate descent}: cycle over coordinates, line-search each on $5$
evenly spaced values in the same box with the others fixed, same objective
and budget. Selection in both uses no labels, and held-out accuracy is computed
once for the selected point. \emph{AdaMerging} (standalone): entropy
descent (Adam, learning rate $10^{-3}$, batch $64$, initialization $0.3$) on
the same unlabeled probe pool as LR-Merge. Costs exclude single-task
fine-tuning, shared by every method.

\paragraph{Forecast experiments.} Pairwise forecasts
(Tables~\ref{tab:e1}, \ref{tab:anchors}) evaluate
$\alpha\in\{0.1,0.2,0.3,0.4,0.5,0.7,1.0\}$ on all $30$ ordered pairs, each
probe seed redrawing the probe and evaluation splits on the same
checkpoints. Full-merge forecasts (Table~\ref{tab:magf}) take the merged
model $\theta_M$ of each base merger, set $\delta=\theta_M-\theta_A$ for each
receiver, and compare $sJ_A(\theta_A)\delta$ with
$z_A(\theta_A+s\delta)-z_A(\theta_A)$, and one JVP covers every $s$, since the
forecast is linear in $s$. The two-JVP column uses the midpoint rule at
$s\in\{1/4,3/4\}$.

\end{document}